\documentclass[11pt]{article}

\usepackage[final]{acl}

\usepackage{times}
\usepackage{latexsym}
\usepackage[rgb]{xcolor} 
\usepackage{rotating}  
\usepackage{booktabs}  
\usepackage{multirow}  
\usepackage{array}     
\usepackage{ragged2e}  
\usepackage{caption}   
\usepackage{threeparttable} 
\usepackage{adjustbox}  
\usepackage{longtable}

\usepackage[T1]{fontenc}

\usepackage[utf8]{inputenc}

\usepackage{microtype}

\usepackage{inconsolata}

\usepackage{graphicx}
\usepackage{subcaption}
\usepackage{tabularx}
\usepackage{booktabs}
\usepackage{amsmath}
\usepackage{amssymb} 
\usepackage{multirow}
\usepackage{makecell}
\usepackage{rotating}   
\usepackage{array}      
\usepackage{seqsplit}

\title{PsyScore: A Psychometrically-Aware Framework for Trait-Adaptive Essay Scoring and ZPD-Scaffolded Feedback}

\author{
 \textbf{Wei Xia\textsuperscript{1}\thanks{\quad Co-first authors with equal contribution.}},
 \textbf{Jin Wu\textsuperscript{2,3}\footnotemark[1]},
 \textbf{Haoran Shi\textsuperscript{1}},
 \textbf{Xiangyu Wang\textsuperscript{1}},
 \textbf{Chanjin Zheng\textsuperscript{2}\thanks{\quad Corresponding author.}}
\\
 \textsuperscript{1}Department of Educational Psychology, East China Normal University,
 \\
 \textsuperscript{2}Shanghai Institute of Artificial Intelligence for Education, East China Normal University,
 \\
 \textsuperscript{3}School of Computer Science and Technology, East China Normal University
\\
 \small{
  \texttt{ \{51264118006, 52275901018\}@stu.ecnu.edu.cn, \href{chjzheng@dep.ecnu.edu.cn}{chjzheng@dep.ecnu.edu.cn}}
 }
}

\begin{document}
\maketitle

\begin{abstract}
Effective Automated Essay Scoring (AES) are expected to support both reliable assessment and actionable instructional feedback. However, existing approaches often treat scoring and feedback as separate components: neural scoring models provide limited interpretability, while Large Language Model (LLM)-based feedback is typically insensitive to learners’ proficiency levels. To address this fragmentation, this work proposes \textbf{PsyScore}, a psychometrically-aware framework that integrates diagnostic assessment with instructional scaffolding through a shared latent ability representation. PsyScore comprises three key modules: a \textbf{Trait-Adaptive Neural IRT Scorer} that incorporates the Graded Partial Credit Model (GPCM) into a neural architecture, enabling the precise estimation of student ability while maintaining psychometric interpretability, a \textbf{ZPD-Scaffolded Feedback Generator}, which conditions multi-agent feedback strategies on the diagnosed ability parameter to adapt instructional focus across different proficiency levels, and a \textbf{Multi-Perspective Feedback Evaluation Strategy} that assesses feedback quality via pairwise preference judgments and student revision simulations. Experiments on the ASAP++ dataset demonstrate that PsyScore achieves competitive scoring performance while providing more pedagogically aligned feedback. 

\end{abstract}

\begin{figure}[t]
    \centering
    \includegraphics[width=0.90\columnwidth]{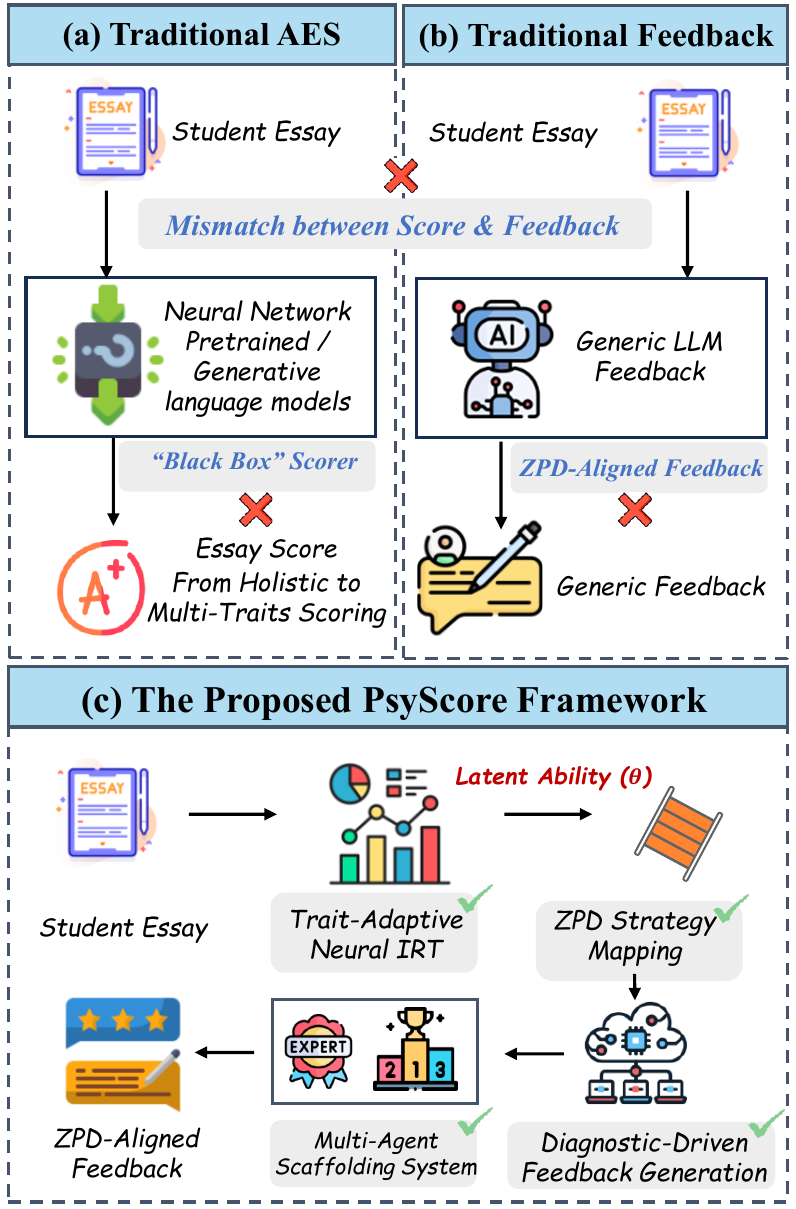}
    \caption{Comparison of traditional approaches and our proposed PsyScore framework. PsyScore aligns diagnostic scoring with scaffolded feedback.}
    \label{fig:introduction}
\end{figure}

\section{Introduction}

Automated Essay Scoring (AES) has become a central component of scalable digital education, evolving from early feature-engineering approaches~\cite{elmassry2025systematic} to deep representation learning and pre-trained language models (PLMs)~\cite{taghipourNeuralApproachAutomated2016a, dong2017, faseeh2024hybrid, kumar2020explainable}. Beyond predictive accuracy, an effective assessment system is expected to operate as a formative loop, in which reliable measurement directly informs targeted instructional scaffolding~\cite{clark2012formative, frank2018formative, WANG2020100449, adair2024ai}. With the rapid expansion of online learning platforms~\cite{mashau2021use, li2025comparative}, the demand for timely, individualized, and actionable feedback has increased substantially. 

Despite the strong performance of recent AES models \cite{ludwig2021automated, elmassry2025systematic}, current solutions often treat diagnostic scoring and pedagogical feedback as separate components \cite{huang2025early}. While recent pioneering efforts such as IFlyEA and CEAES \cite{gong2021iflyea, li2025ceaes} have begun to bridge this gap by modeling the synergy between scoring and review generation, these approaches typically lack a principled psychometric grounding. This fragmentation gives rise to three persistent limitations (Figure~\ref{fig:introduction}). First, most neural AES models function as opaque predictors \cite{misgna2024survey}. Although they optimize standard objectives such as mean squared error \cite{ferrara2022validity}, they are weakly grounded in educational measurement theory~\cite{voss2025comparison}, raising concerns about psychometric validity, interpretability, and fairness \cite{schaller2024fairness, shin2021automated, Jiang2023Interpretability-Aware}. Second, the separation between scoring and feedback restricts diagnostic utility. A single holistic score or even loosely coupled multi-trait predictions, fails to capture the structured nature of writing proficiency needed to support targeted instructional decisions \cite{ono2019holistic, imbler2022teaching, chen2023multistrategy, stahl2024exploring}. Third, while Large Language Models (LLMs) are capable of generating fluent feedback \cite{policar2025automated}, such feedback is typically ability-agnostic. Without explicit modeling of learner proficiency or the Zone of Proximal Development (ZPD) \cite{chaiklin2003zone}, LLM-generated comments frequently exhibit a cognitive mismatch, being overly procedural for advanced learners or overly abstract for novices \cite{jacobsen2025promises, yang2025does}.

\begin{figure*}[t] 
    \centering
    \includegraphics[width=1.0\textwidth]{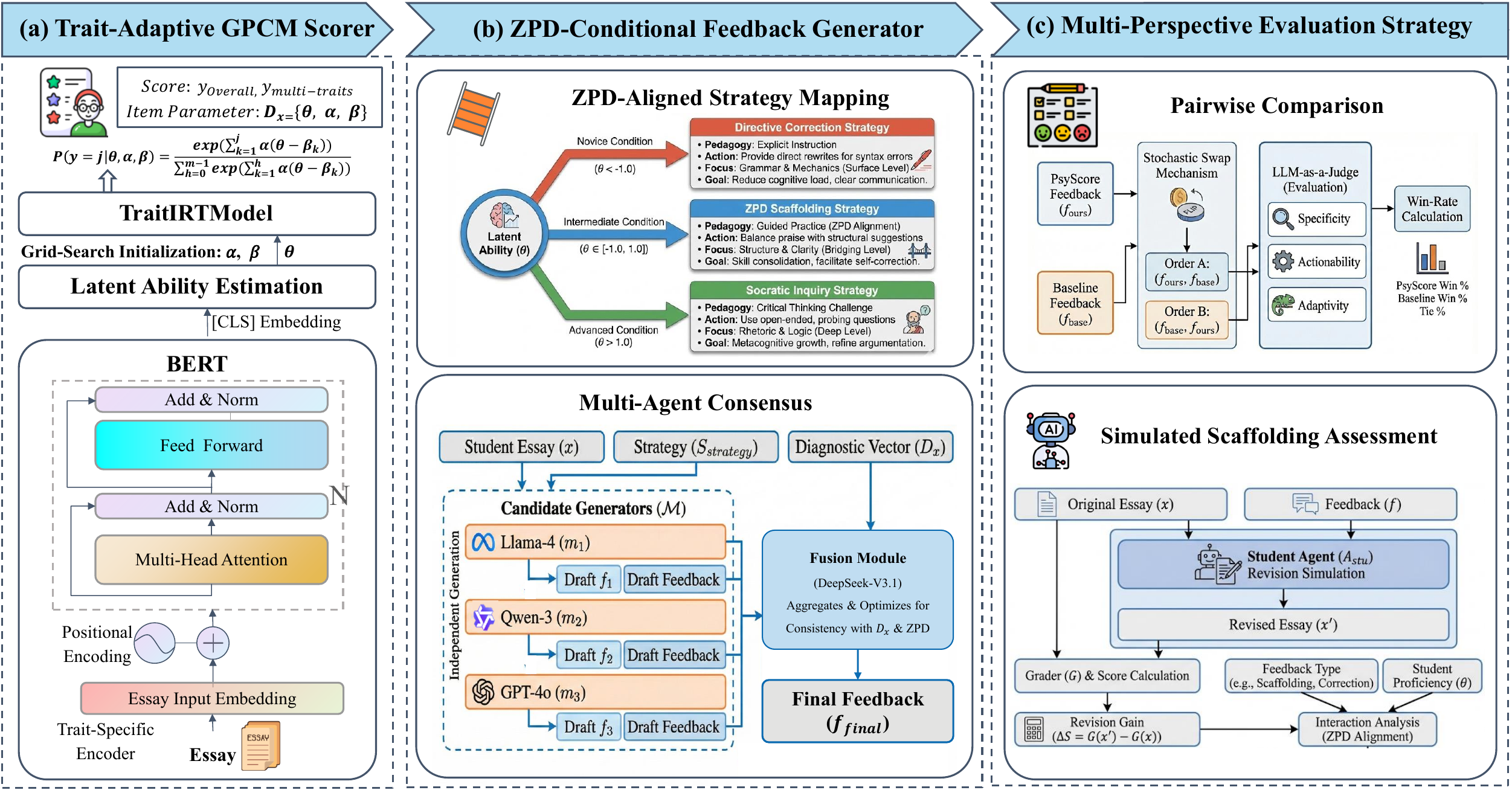} 
    \caption{Overview of the PsyScore framework. 
    (a) \textbf{Trait-Adaptive GPCM Scorer} estimates the student's latent ability ($\theta$) and outputs a diagnostic vector ($D_x$). 
    (b) \textbf{ZPD-Conditional Feedback Generator} synthesizes consensus feedback ($f_{final}$) by mapping $\theta$ to adaptive strategies via multi-agent fusion. 
    (c) \textbf{Multi-Perspective Evaluation} validates quality via intrinsic LLM-based comparison and extrinsic simulated revision.}  
    \label{fig:framework}
\end{figure*}

These limitations suggest a shared underlying cause: the absence of the principled latent representation that simultaneously supports measurement and instruction. Motivated by this observation, we explore the hypothesis that modeling student ability within a shared psychometric latent space can unify interpretable scoring and ZPD-aligned feedback. To this end, we propose \textbf{PsyScore}, a psychometrically-aware framework that integrates diagnostic assessment with instructional scaffolding through a latent representation (Figure~\ref{fig:framework}).

PsyScore comprises two tightly coupled components. On the assessment side, it introduces a \textbf{Trait-Adaptive Neural IRT Scorer} that embeds Item Response Theory into a neural architecture to estimate interpretable student ability across multiple writing traits. This latent ability representation is then shared with a \textbf{ZPD-Scaffolded Feedback Generator}, which conditions pedagogical strategies on the diagnosed proficiency level within a multi-agent framework, enabling feedback that is aligned with learners’ cognitive readiness.

The contributions of this work are threefold:
\begin{itemize}
\setlength{\itemsep}{0pt}
\item \textbf{Psychometric Calibration.} We propose a trait-adaptive initialization strategy for neural IRT models that aligns discrimination and difficulty parameters with psychometric priors.
\item \textbf{Pedagogical Alignment.} We introduce a ZPD-scaffolded multi-agent feedback framework that conditions instructional strategies on diagnosed learner ability ($\theta$).
\item \textbf{Empirical Insight.} We conduct a dual-layer evaluation that analyzes the relationship between scoring performance and downstream learning outcomes across proficiency levels.
\end{itemize}

\section{Related Work}

\subsection{From Holistic Scoring to Multi-Trait Analysis}
AES research has evolved from early feature-engineering paradigms \cite{chenAutomatedEssayScoring2014a} to modern deep neural networks \cite{taghipourNeuralApproachAutomated2016a} and pre-trained language models \cite{ludwigAutomatedEssayScoring2021}. While achieving high predictive accuracy, holistic scoring offers limited diagnostic value. To address this, recent frameworks like ArTS \cite{do-etal-2024-autoregressive-multi} and T-MES \cite{wang-liu-2025-mes} adopt multi-task learning to evaluate essays along multiple dimensions. However, these models remain primarily predictive: trait scores are treated as parallel outputs rather than manifestations of a shared latent ability. This lack of psychometric grounding limits their interpretability and their capacity to guide principled instructional feedback.

\subsection{Item Response Theory in AES}
Item Response Theory (IRT) provides a probabilistic framework linking latent ability to observed performance \cite{LI2025103913,samejima1969estimation, utoItemResponseTheory2018,wu2025metacdmetalearningframework}. Within AES, recent neuro-symbolic approaches have integrated IRT to enhance interpretability and fairness \cite{shin2021automated, Jiang2023Interpretability-Aware}. Notably, \citet{shibataAnalyticAutomatedEssay2022} combined IRT with deep learning for analytic scoring, achieving robust multidimensional assessment. However, these prior neural IRT methods focus predominantly on score estimation as a regularization term or multi-task objective. They typically suffer from calibration instability and, crucially, fail to utilize the estimated latent ability ($\theta$) as a control signal for downstream tasks, leaving the critical loop between diagnosis and instructional intervention unconnected.
While recent work such as SMART \cite{scarlatos2025smart} uses IRT-aligned simulated students primarily for item difficulty prediction, PsyScore framework extends the application of IRT latent traits to individualized diagnostic feedback generation, focusing on maximizing the pedagogical gain within the student's ZPD.

\subsection{LLM-based Personalized Feedback}

The transition from Seq2Seq models \cite{LIU2024123043} to Large Language Models (LLMs) has enabled scalable, fluent feedback generation \cite{stahl-etal-2024-exploring}. However, current prompting strategies often lack explicit adaptation mechanisms, risking a cognitive mismatch where feedback is either too simplistic or overly complex for the learner. While benchmarks like PROF \cite{nair-etal-2024-closing} and eRevise \cite{liu-etal-2025-erevise} assess feedback utility based on revision outcomes, they do not verify cognitive alignment. Consequently, existing systems often suffer from content homogenization \cite{ICLR2024_02dec887} and fail to operationalize Zone of Proximal Development principles, limiting their efficacy as adaptive instructional scaffolds.

\section{The PsyScore Framework}
\subsection{Overall Architecture}
PsyScore establishes an integrated framework that couples psychometrically grounded latent trait analysis with generative instructional scaffolding. As illustrated in Figure~\ref{fig:framework}, the system pipeline consists of three mathematically integrated modules: 
(1) A \textbf{Trait-Adaptive Neural GPCM Scorer} for disentangled representation learning and latent parameter estimation. 
(2) A \textbf{ZPD-Conditional Feedback Generator} utilizing a ``Generate-and-Fuse'' mechanism with heterogeneous agents. 
And (3) a \textbf{Multi-Perspective Feedback Evaluation Strategy} quantifying both semantic alignment and pedagogical efficacy.

\subsection{Trait-Adaptive Neural GPCM Scorer}
This module functions as a probabilistic regressor that maps input text $x$ to a psychometric latent space, which is a common strategy to improve robustness in high-stakes educational assessment.

\subsubsection{Latent Ability Estimation}
For a given writing trait $t$, a BERT-based encoder extracts a dense contextual representation $h_x \in \mathbb{R}^{d}$. This vector is projected onto a scalar latent ability $\theta$ via a linear transformation. To align with the standard normal assumption of IRT ($\theta \sim \mathcal{N}(0, 1)$) and ensure numerical stability within the GPCM exponent, we enforce a hard clipping constraint:
\begin{equation}
\small
\theta = \text{Clamp}(W_\theta h_x + b_\theta, \min=-3.0, \max=3.0)
\end{equation}
This constraint restricts $\theta$ to the effective discrimination interval $[-3, 3]$, preventing gradient explosion during end-to-end training.

\subsubsection{GPCM Probability Modeling}
The constrained $\theta$ serves as the input to a Graded Partial Credit Model (GPCM) layer. The probability of assigning a discrete score category $k \in \{0, \dots, K\}$ is modeled as:
\begin{equation}
\small
P(y=k | \theta, a_t, \mathbf{b}_t) = \frac{\exp \left( \sum_{j=0}^{k} a_t (\theta - b_{t,j}) \right)}{\sum_{c=0}^{K} \exp \left( \sum_{j=0}^{c} a_t (\theta - b_{t,j}) \right)}
\end{equation}
where $a_t \in \mathbb{R}^+$ is the learnable discrimination parameter, and $\mathbf{b}_t = \{b_{t,0}, \dots, b_{t,K}\}$ represents the step difficulty thresholds. The model is optimized by minimizing the Negative Log-Likelihood (NLL) of the true labels $y_{true}$:
\begin{equation}
\mathcal{L} = -\sum_{i=1}^{N} \log P(y_i | \theta_i, a_t, \mathbf{b}_t)
\end{equation}

\subsubsection{Initialization and Ensemble Strategy}
To resolve the non-convexity of the GPCM loss landscape, which has been noted as a practical challenge in neural IRT-based models, this work employ a grid-search initialization strategy. We pre-compute the optimal priors for $a_{init}$ and $b_{init}$ over the hyperparameter space $\mathcal{H} = \{0.5, 1.0, 1.5\} \times \mathbb{R}^{K}$ to prevent mode collapse. During inference, the system aggregates parameters from all $k$ folds to construct a robust diagnostic vector $D_x = \{\bar{\theta}, \bar{a}, \bar{\mathbf{b}}\}$, which serves as the conditioning signal for the feedback module.

\subsection{ZPD-Conditional Feedback Generator}
This module operates as a conditional generation pipeline $F(x, D_x) \rightarrow Y$. It leverages diagnostic parameters to synthesize feedback via a multi-agent fusion architecture.

\subsubsection{ZPD-Aligned Strategy Mapping}
We define a deterministic mapping function $M: \mathbb{R} \rightarrow \mathcal{S}$ inspired by the notion of the ZPD, that translates the estimated ability $\bar{\theta}$ into a pedagogical control token:
\begin{equation}
\small
S_{strategy} = 
\begin{cases} 
\text{Explicit Correction} & \text{if } \bar{\theta} < -1.0 \\
\text{Scaffolding} & \text{if } \bar{\theta} \in [-1.0, 1.0] \\
\text{Socratic Questioning} & \text{if } \bar{\theta} > 1.0 
\end{cases}
\end{equation}
Additionally, traits with high discrimination ($\bar{a} > 1.2$) are identified as ``High-Information Dimensions'' and are assigned higher weights in the prompt attention mechanism. This threshold is grounded in standard psychometric evaluation criteria, where discrimination values above 1.2 are categorized as ``High'' to ``Very High'', ensuring the selected traits possess sufficient statistical validity to distinguish between proficiency levels \cite{baker2001basics}. The ZPD boundary values are designed to operationalize scaffolding theory within a computational framework \cite{wood1976role}.

\subsubsection{Multi-Agent Consensus and Debiasing}
To maximize pedagogical diversity, consistent with prior findings on diversity-oriented generation, we deploy a set of heterogeneous LLMs $\mathcal{M} = \{\texttt{\scriptsize Llama-4-Scout}, \texttt{\scriptsize Qwen3-235B-A22B-Instruct-2507}, \texttt{\scriptsize GPT-4o}\}$ as candidate generators. Each model $m \in \mathcal{M}$ independently generates a draft $f_m$ conditioned on the essay $x$ and strategy $S_{strategy}$.
A superior model (\texttt{DeepSeek-V3.1}) functions as the Fusion Module. It aggregates the candidate set $\{f_m\}_{m=1}^{|\mathcal{M}|}$ into a final response $f_{final}$ by optimizing for consistency with the diagnostic vector $D_x$:
\begin{equation}
f_{final} = \text{Fusion}(\{f_1, f_2, f_3\}, D_x)
\end{equation}
This fusion step resolves semantic conflicts and ensures the linguistic complexity of the output strictly adheres to the learner's ZPD.

\subsection{Multi-Perspective Feedback Evaluation Strategy}
Evaluating open-ended instructional feedback requires moving beyond surface-level lexical overlap. We therefore establish a three-tier evaluation framework to holistically assess the generated feedback: 
\textbf{Preference-based Comparison}: Pairwise ranking by independent LLM judges to measure relative pedagogical quality.
\textbf{Simulated Revision}: Measuring the normalized score gain achieved when a simulated student agent revises essays conditioned on the feedback.
\textbf{Human Expert Evaluation}: Fine-grained assessment by education professionals along theoretically grounded dimensions.

\subsubsection{Preference Comparison Evaluation with LLM Judges}
To mitigate the inherent calibration biases of absolute scoring, we adopt a pairwise preference comparison paradigm. This approach compels the judge model to perform discriminative analysis between feedback generated by \text{PsyScore-AEF} (The PsyScore Automated Essay Feedback module) and baseline models, effectively capturing fine-grained qualitative differences.

This work designated \texttt{GPT-5} and \texttt{Gemini-3-pro} as independent judges to perform pairwise comparisons, selecting them for their superior discriminative performance. The evaluation process adheres to two rigorous protocols: (1) Anonymized Evaluation, a double-blind setup where models receive de-identified texts labeled solely as Response A and B; and (2) Positional Bias Mitigation, which employs stochastic position swapping to counteract the ``lead bias'' prevalent in LLMs by evaluating each pair twice in randomized orders to ensure consistent judgment.

Judges assign a ``Win'',``Tie'', or ``Loss'' across multiple pedagogical dimensions. We report the aggregate win rate to quantify the relative superiority of \text{PsyScore-AEF} in generating personalized, high-quality feedback.

\subsubsection{Simulation-based Revision Assessment}
To evaluate whether feedback provides cognitive scaffolding rather than direct answers, we implement a simulation-based revision protocol. We construct a ``Student Agent'' $A_{stu}$ conditioned on ability-specific profiles aligned with the estimated latent trait $\theta$. The agent is instructed to emulate a learner at the diagnosed proficiency level and revise the original essay $x$ based strictly on the received feedback $f$, yielding a revised version $x'$.

To ensure measurement invariance across the draft-revision sequence, we utilize the fine-tuned PsyScore-AES as the uniform scoring function $S(\cdot)$. Given the heterogeneous score ranges across prompts (e.g., 0--3 vs. 0--60), absolute differences $\Delta S = S(x') - S(x) $ would introduce range-dependent bias. We therefore compute a normalized revision gain $\Delta S_{\text{norm}}$ to project improvements onto a unified $[-1, 1]$ scale:
\begin{equation}
\Delta S_{\text{norm}} = \frac{S(x') - S(x)}{S_{\text{max}} - S_{\text{min}}}
\end{equation}
where $S_{\text{max}}$ and $S_{\text{min}}$ denote the prompt-specific score bounds. This metric quantifies the marginal pedagogical contribution of the feedback while eliminating confounding effects from disparate scoring scales.

\subsubsection{Human Expert Evaluation}
To validate the pedagogical efficacy of the generated feedback and to mitigate potential biases inherent in LLM-as-a-judge evaluations, we conducted a double-blind study involving three senior education experts. The experts assessed a stratified random sample of 80 essays along with their corresponding feedback across five theoretically grounded dimensions. The detailed rubric defining each dimension is provided in Appendix~\ref{sec:appendix-f}. 

\begin{table*}[t]
    \centering
    \caption{Main results on each prompt. The average QWK across all traits for each prompt is reported.}
    \label{tab:AES_results_by_prompt}
    \setlength{\tabcolsep}{4pt} 

    \begin{tabular}{lccccccccc}
        \toprule
        \textbf{Model} & \textbf{P1} & \textbf{P2} & \textbf{P3} & \textbf{P4} & \textbf{P5} & \textbf{P6} & \textbf{P7} & \textbf{P8} & \textbf{AVG} \\
        \midrule
        STL-LSTM \cite{dong2017} & 0.690 & 0.622 & 0.663 & 0.719 & 0.719 & 0.753 & 0.704 & 0.592 & 0.683 \\
        HISK \cite{cozma-etal-2018-automated} & 0.674 & 0.586 & 0.651 & 0.681 & 0.693 & 0.709 & 0.641 & 0.516 & 0.644 \\
        MTL-BiLSTM \cite{kumar2022many} & 0.670 & 0.611 & 0.647 & 0.708 & 0.704 & 0.712 & 0.684 & 0.581 & 0.665 \\
        DualTrans \cite{choDualscaleBERTUsing2024a} & 0.712 & 0.671 & 0.690 & 0.760 & 0.714 & 0.740 & 0.748 & 0.620 & 0.707 \\
        ArTS \cite{do-etal-2024-autoregressive} & 0.708 & 0.706 & 0.704 & 0.767 & 0.723 & \textbf{0.776} & 0.749 & 0.603 & 0.717 \\
        SaMRL-large \cite{do-etal-2024-autoregressive-multi} & 0.702 & \underline{0.711} & \underline{0.708} & 0.766 & 0.722 & \underline{0.773} & 0.743 & \underline{0.649} & \underline{0.722} \\
        T-MES \cite{wang-liu-2025-mes} & \underline{0.728} & 0.684 & 0.702 & \textbf{0.771} & \underline{0.726} & 0.754 & \underline{0.755} & 0.629 & 0.719 \\
        \citet{shibataAnalyticAutomatedEssay2022} & 0.667 & 0.642 & 0.655 & 0.669 & 0.658 & 0.651 & 0.644 & 0.638 & 0.653 \\
        \midrule
        \textbf{PsyScore-AES (Our Model)} & \textbf{0.744} & \textbf{0.723} & \textbf{0.732} & \underline{0.770} & \textbf{0.750} & \underline{0.773} & \textbf{0.760} & \textbf{0.730} & \textbf{0.747} \\
        PsyScore-AES (w/o-IRT) & 0.709 & 0.685 & 0.707 & 0.756 & 0.714 & 0.737 & 0.717 & 0.613 & 0.705 \\
        \bottomrule
    \end{tabular}
    \vspace{0.9em}
    \begin{minipage}{\textwidth}
    \par\raggedright\noindent\footnotesize
    \textit{\textbf{w/o IRT: }This denotes the ablation experiment in which the IRT module is removed from the proposed PsyScore-AES model.}
    \end{minipage}
\end{table*}

\begin{table*}[t]
    \centering
    \caption{Main results of each trait. The average QWK across all prompts for each trait is reported.}
    \label{tab:AES_results_by_trait_non_rotated}
    \setlength{\tabcolsep}{5pt} 

    \resizebox{\textwidth}{!}{%
        \begin{tabular}{lcccccccccccc}
            \toprule
            \textbf{Model} & \textbf{Ovr.} & \textbf{Cont.} & \textbf{PA} & \textbf{Lang.} & \textbf{Nar.} & \textbf{Org.} & \textbf{Conv.} & \textbf{WC} & \textbf{SF} & \textbf{Style} & \textbf{Voice} & \textbf{AVG} \\
            \midrule
            STL-LSTM & 0.750 & 0.707 & 0.731 & 0.640 & 0.699 & 0.649 & 0.605 & 0.621 & 0.612 & 0.659 & 0.544 & 0.656 \\
            HISK & 0.718 & 0.679 & 0.697 & 0.605 & 0.659 & 0.610 & 0.527 & 0.579 & 0.553 & 0.609 & 0.489 & 0.611 \\
            MTL-BiLSTM & 0.762 & 0.719 & 0.731 & 0.659 & 0.703 & 0.669 & 0.656 & 0.676 & 0.625 & \underline{0.693} & 0.610 & 0.682 \\
            DualTrans & 0.764 & 0.685 & 0.701 & 0.604 & 0.668 & 0.615 & 0.560 & 0.615 & 0.598 & 0.632 & 0.582 & 0.639 \\
            ArTS & \underline{0.778} & 0.726 & 0.732 & 0.660 & 0.704 & 0.682 & 0.668 & 0.674 & 0.663 & 0.689 & \underline{0.619} & 0.690 \\
            SaMRL-large & 0.774 & \underline{0.730} & 0.750 & \underline{0.702} & \underline{0.730} & \underline{0.685} & \underline{0.686} & \underline{0.679} & 0.675 & \underline{0.693} & 0.590 & \underline{0.699} \\
            T-MES & 0.754 & \underline{0.730} & \underline{0.751} & 0.698 & 0.725 & 0.672 & 0.668 & \underline{0.679} & \underline{0.678} & \textbf{0.721} & 0.570 & 0.695 \\
            \midrule
            \textbf{PsyScore-AES (Our)} & \textbf{0.790} & \textbf{0.761} & \textbf{0.762} & \textbf{0.708} & \textbf{0.749} & \textbf{0.707} & \textbf{0.733} & \textbf{0.725} & \textbf{0.725} & 0.683 & \textbf{0.740} & \textbf{0.735} \\
            PsyScore-AES (w/o-IRT) & 0.735 & 0.739 & 0.726 & 0.676 & 0.717 & 0.650 & 0.671 & 0.680 & 0.653 & 0.619 & 0.636 & 0.682 \\
            \bottomrule
        \end{tabular}%
    }
\end{table*}

\begin{table}[t]
    \centering
    \caption{Optimal psychometric initialization parameters derived via grid search. Distinct configurations validate the necessity of trait-adaptive modeling.}
    \label{tab:optimal_hyperparams}
    \setlength{\tabcolsep}{8pt}
    \normalsize
    \begin{tabular}{lcc}
        \toprule
        \textbf{Trait} & \boldmath$a_{init}$ & \boldmath$b_{range}$ \\
        \midrule
        Overall & 0.5 & $[-3, 3]$ \\
        Content & 1.5 & $[-2, 2]$ \\
        Organization & 0.5 & $[-3, 3]$ \\
        Word Choice & 1.0 & $[-3, 3]$ \\
        Sentence Fluency & 1.0 & $[-1, 1]$ \\
        Conventions & 1.5 & $[-1, 1]$ \\
        Prompt Adherence & 1.5 & $[-2, 2]$ \\
        Narrativity & 1.5 & $[-1, 1]$ \\
        Language & 0.5 & $[-3, 3]$ \\
        Style & 1.5 & $[-1, 1]$ \\
        Voice & 0.5 & $[-2, 2]$ \\
        \bottomrule
    \end{tabular}
\end{table}

\begin{figure*}[t] 
    \centering
    \label{fig:preference-evaluation}
    \begin{subfigure}[b]{0.48\textwidth}
        \centering
        \includegraphics[width=\linewidth]{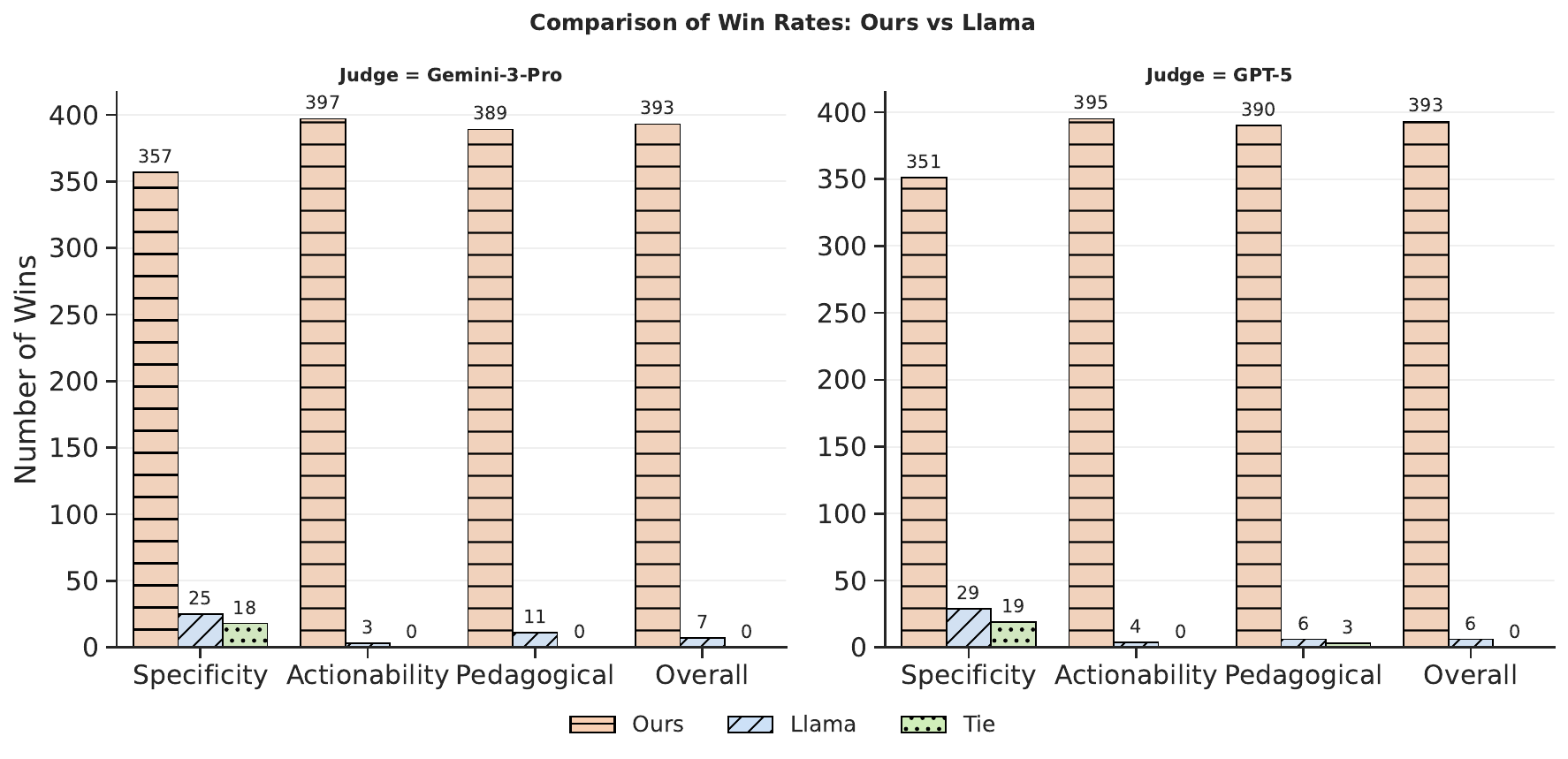} 
        \caption{PsyScore-AEF vs. Llama-4-Scout}
        \label{fig:vs_llama}
    \end{subfigure}
    \hfill
    \begin{subfigure}[b]{0.48\textwidth}
        \centering
        \includegraphics[width=\linewidth]{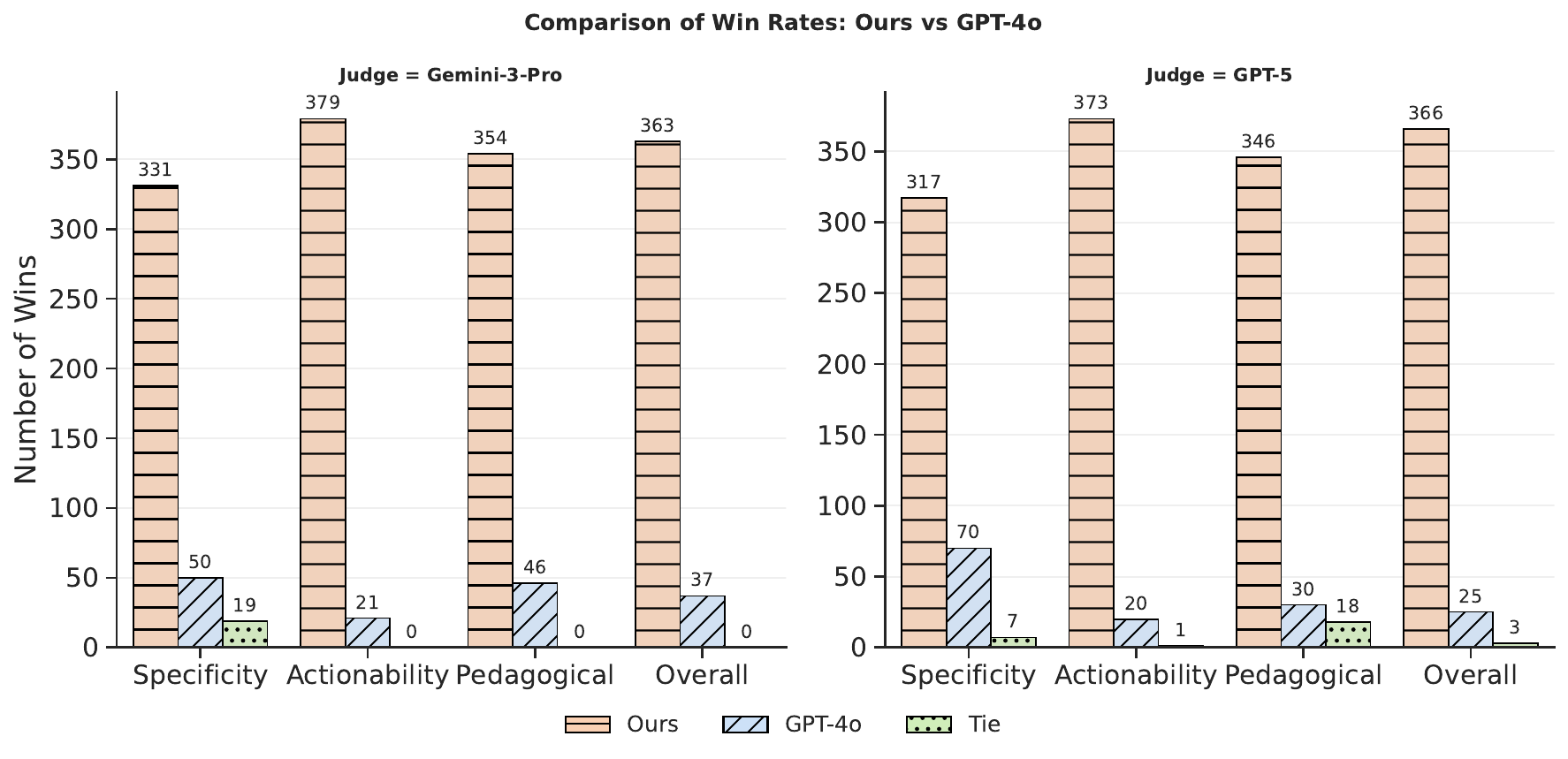}
        \caption{PsyScore-AEF vs. GPT-4o}
        \label{fig:vs_gpt4o}
    \end{subfigure}

    \vspace{1em} 
    \begin{subfigure}[b]{0.48\textwidth}
        \centering
        \includegraphics[width=\linewidth]{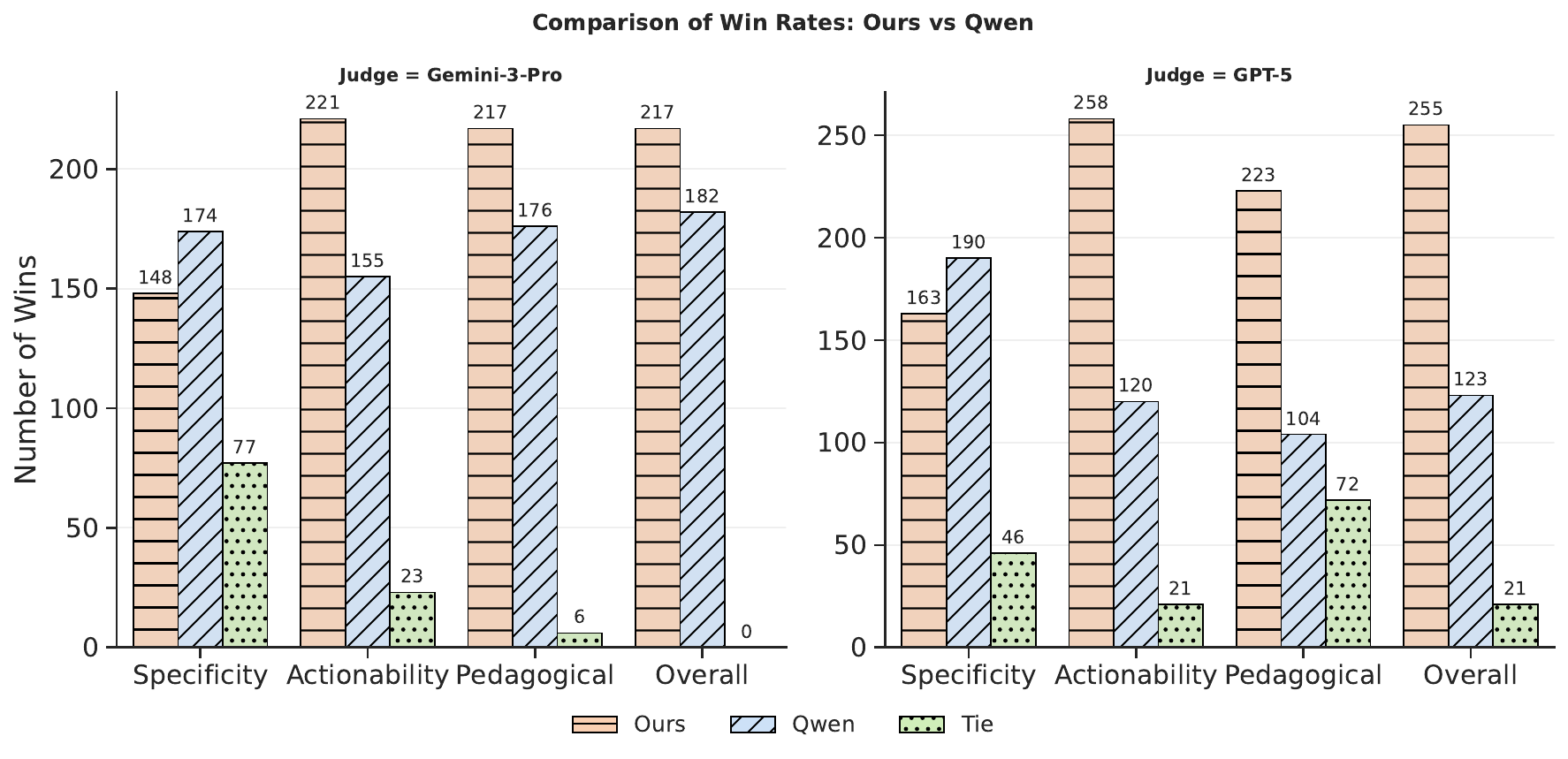}
        \caption{PsyScore-AEF vs. Qwen3-235B-A22B-Instruct-2507}
        \label{fig:vs_qwen}
    \end{subfigure}
    \hfill
    \begin{subfigure}[b]{0.48\textwidth}
        \centering
        \includegraphics[width=\linewidth]{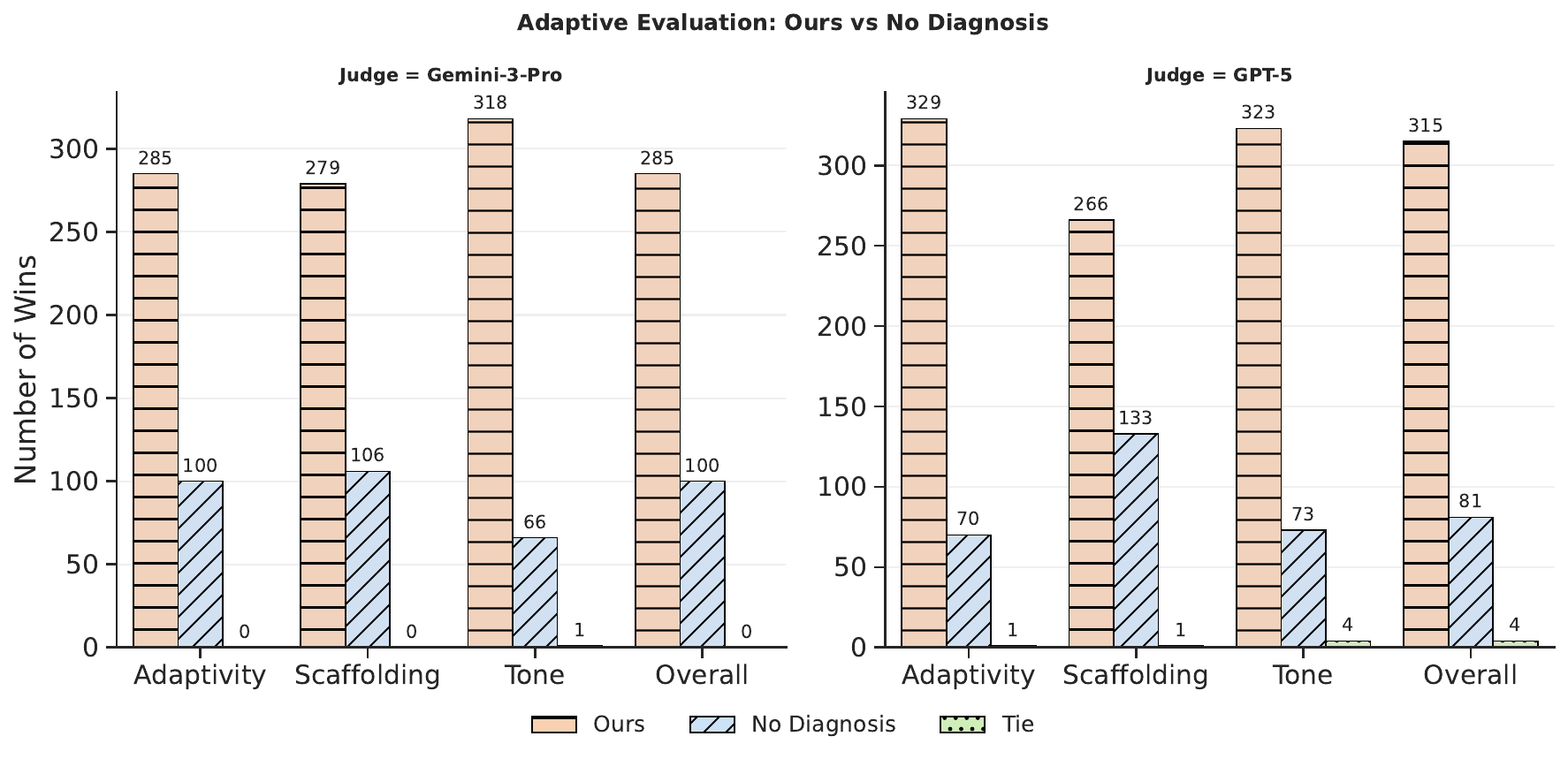}
        \caption{PsyScore-AEF vs. No IRT}
        \label{fig:vs_noirt}
    \end{subfigure}

    \caption{\textbf{Pairwise preference evaluation results across four baselines.} The bars represent the number of wins awarded by judges. PsyScore demonstrates consistent superiority across both open-source (a-c) and closed-source (d) models, particularly in \textit{Actionability} and \textit{Adaptivity}.}
    \label{fig:ablation_feedback}
\end{figure*}

\section{Experiments and Results}

\subsection{Dataset}
This work conduct experiments on the \textbf{ASAP++} dataset~\cite{mathias2018asap++}, utilizing a standard 6:2:2 data split (60\% training, 20\% validation, and 20\% testing). This dataset extends the original ASAP corpus by providing trait-specific scores (e.g., Content, Organization,Conventions,Word Choice,Sentence Fluency) alongside holistic scores, enabling granular psychometric analysis.


\subsection{Implementation Setup}
All experiments were implemented using \texttt{PyTorch} and the \texttt{bert-base-uncased}\cite{devlin-etal-2019-bert}. To ensure reproducibility, we fixed the random seed to 42 for all data splitting and model initialization steps. The computational processes were executed on a single NVIDIA RTX A6000 GPU (48 GB of VRAM). The experiments leveraged CUDA 12.1 and cuDNN 8.9 to optimize GPU performance, with all operations carried out on a Linux-based environment (Ubuntu 22.04) to ensure reproducibility and stability of results.

\subsubsection{Scoring Model Training}
For the PsyScore-AES module, we fine-tune the \texttt{bert-base-uncased} backbone using a 5-fold cross-validation strategy; final results are reported as the ensemble average to minimize variance. The model is optimized with AdamW (learning rate 5e-6, batch size 16) for up to 30 epochs. We apply early stopping with a patience of 10 epochs, monitoring the Quadratic Weighted Kappa (QWK) on the validation set to prevent overfitting.

For the psychometric calibration, the grid search for GPCM parameters is performed within the following spaces: discrimination prior $a_{init} \in \{0.5, 1.0, 1.5\}$ and difficulty range $b_{range} \in \{[-1, 1], [-2, 2], [-3, 3]\}$. We select the optimal configuration for each trait based on validation performance. The final trait-specific hyperparameters are detailed in Table~\ref{tab:optimal_hyperparams}.

\subsubsection{Feedback Model Configuration}
The PsyScore-AEF module implement a Generate-then-Fuse multi-agent system where the latent trait $\theta$ serves as a cognitive control signal. First, three heterogeneous agents (\texttt{Llama-4-Scout}, \texttt{\seqsplit{Qwen3-235B-A22B-Instruct-2507}}, \texttt{GPT-4o}) generate initial drafts to ensure broad pedagogical coverage within the learner's ZPD. Subsequently, an expert agent (\texttt{DeepSeek-V3.1}) synthesizes these drafts into a coherent response, filtering hallucinations and aligning the instructional scaffolding with the diagnosed ability $\theta$.

\subsubsection{Sampling Strategy and Data Validity}

To balance evaluation depth with computational efficiency, we conducted the feedback analysis on a stratified random sample of $N=400$ essays (50 per prompt) from the ASAP++ dataset. This subset preserves the original distribution of writing genres—narrative, persuasive, and expository—ensuring statistical representativeness. A negligible portion of cases ($<1\%$) was excluded due to stochastic API invocation issues. Given this minimal attrition rate, the exclusions do not compromise the statistical validity of the results.

\subsection{Main Results: Scoring Performance}
Tables~\ref{tab:AES_results_by_prompt} and \ref{tab:AES_results_by_trait_non_rotated} summarize the comparative scoring performance of PsyScore-AES against state-of-the-art baselines.

\subsubsection{Prompt-level Comparison} 
PsyScore-AES achieves a new state-of-the-art average Quadratic Weighted Kappa (QWK) of 0.747, surpassing the strongest baseline (SaMRL-large, 0.722) and ranking first in 6 out of 8 prompts. On the narrative-focused Prompt 8, our model attains a QWK of 0.730, compared with the previous best of 0.649. This improvement suggests that the Trait-Adaptive Calibration strategy enhances model performance on prompts with higher narrative content.

\subsubsection{Trait-level Assessment} 
At the trait level, PsyScore-AES ranks first in 10 out of 11 evaluated dimensions, achieving a mean QWK of 0.735. In the Voice dimension, which is typically more subjective, the model improves the QWK from 0.619 (ArTS) to 0.740. While slightly trailing T-MES in the \textit{Style} dimension (0.683 vs. 0.721), PsyScore-AES maintains the highest overall performance across traits, indicating the effectiveness of disentangled representation learning in capturing multiple aspects of essay quality.

\subsubsection{Ablation Analysis} 
We evaluate the contribution of the psychometric calibration by comparing the full model with a variant in which the IRT module is included without grid-search initialization of discrimination ($a$) and difficulty ($b$) parameters. Removing this initialization results in a performance drop from QWK 0.747 to 0.705. This observation highlights the importance of properly optimizing IRT priors to align neural representations with expected psychometric distributions.

\subsection{Statistical Significance Analysis}
We assess significance of QWK differences between PsyScore-MAES and BERT using the \textbf{Wilcoxon Signed-Rank Test} (one-tailed, $n=5$ folds). This non-parametric choice avoids normality assumptions and leverages paired test splits. Under the one-tailed criterion, PsyScore-MAES outperforming BERT on all five folds yields $p = 1/2^5 = 0.03125$, the minimum attainable $p$-value. Full per-trait results with exact $p$-values are provided in Appendix~\ref{app:full_statistical_table}. Across prompts, PsyScore-MAES achieves statistically significant gains on the majority of traits, with the largest improvements on wide-range, fine-grained tasks (e.g., Prompts~7--8) and on core dimensions such as \textit{Content} and \textit{Organization}. Surface-level traits show smaller, often non-significant differences, indicating that the psychometric layer primarily enhances modeling of higher-order writing constructs.

\subsection{Robustness Analysis of Hyperparameter Initialization}

To assess the sensitivity of the PsyScore-MAES framework to IRT initialization parameters, we conducted experiments under nine distinct parameter configurations. The settings spanned discrimination values \( a \in \{0.5, 1.0, 1.5\} \) and difficulty ranges \( b_{range} \in \{[-1,1], [-2,2], [-3,3]\} \). Table~\ref{tab:stability_prompt} and Table~\ref{tab:stability_trait} report the mean Quadratic Weighted Kappa (QWK), standard deviation (\(\sigma\)), and Coefficient of Variation (CV) at the prompt level and trait level, respectively. In psychometric practice, a CV below \(5\%\) is generally regarded as indicative of high stability.

As shown in Table~\ref{tab:stability_prompt}, the Coefficient of Variation (CV) across all prompts ranges from \(0.80\%\) to \(3.27\%\), with an average of \(1.67\%\), substantially below the \(5\%\) stability threshold. This confirms that PsyScore-MAES is highly robust to variations in IRT initialization, maintaining consistent scoring regardless of starting parameter values. Notably, performance and stability are not coupled: the framework adapts reliably across prompts with differing score distributions and task types.

\begin{table}[!h]
\centering
\caption{Performance statistics across different initialization settings (Prompt-level).}
\label{tab:stability_prompt}
\normalsize
\begin{tabular}{lcccc}
\toprule
\textbf{Prompt} & \textbf{Mean QWK} & \textbf{Std. (\(\sigma\))} & \textbf{CV (\%)} \\
\midrule
Prompt 1 & 0.730 & 0.016 & 2.290 \\
Prompt 2 & 0.706 & 0.012 & 1.760 \\
Prompt 3 & 0.714 & 0.014 & 1.970 \\
Prompt 4 & 0.762 & 0.008 & 1.090 \\
Prompt 5 & 0.711 & 0.017 & 2.390 \\
Prompt 6 & 0.746 & 0.024 & 3.270 \\
Prompt 7 & 0.746 & 0.013 & 1.800 \\
Prompt 8 & 0.706 & 0.006 & 0.800 \\
\midrule
\textbf{Average} & \textbf{0.727} & \textbf{0.0121} & \textbf{1.67} \\
\bottomrule
\end{tabular}
\end{table}
Table~\ref{tab:stability_trait} reports trait-level statistics. The \textit{Overall} trait exhibits the lowest CV (\(0.99\%\)), while subjective dimensions such as \textit{Style} and \textit{Voice} show slightly higher variability (\(4.72\%\) and \(4.00\%\), respectively), remaining well within acceptable bounds. These results demonstrate that PsyScore-MAES yields reliable scoring outcomes across all traits, irrespective of initialization settings.
\begin{table}[!h]
\centering
\caption{Performance statistics across different initialization settings (Trait-level).}
\label{tab:stability_trait}
\setlength{\tabcolsep}{1.5pt}
\small
\begin{tabular}{lcccc}
\toprule
\textbf{Trait} & \textbf{Mean QWK} & \textbf{Std. (\(\sigma\))} & \textbf{CV (\%)} \\
\midrule
Overall & 0.777 & 0.008 & 0.990 \\
Content & 0.743 & 0.012 & 1.590 \\
Organization & 0.695 & 0.015 & 2.140 \\
Word Choice & 0.702 & 0.012 & 1.750 \\
Sentence Fluency & 0.700 & 0.011 & 1.530 \\
Conventions & 0.712 & 0.012 & 1.690 \\
Prompt Adherence & 0.739 & 0.017 & 2.280 \\
Narrativity & 0.717 & 0.015 & 2.020 \\
Language & 0.686 & 0.022 & 3.230 \\
Style & 0.666 & 0.032 & 4.720 \\
Voice & 0.712 & 0.029 & 4.000 \\
\midrule
\textbf{Average} & \textbf{0.714} & \textbf{0.012} & \textbf{1.710} \\
\bottomrule
\end{tabular}
\end{table}

\begin{table}[!h]
\centering
\caption{Normalized revision gains and final scores stratified by student proficiency.}
\label{tab:revision-gain}
\setlength{\tabcolsep}{4pt}
\begin{tabular}{lccc}
\hline
\textbf{Group} & \textbf{Count} & \textbf{PsyScore} & \textbf{No-IRT} \\
\hline
Low (\( \theta < -1 \)) & 66 & \textbf{0.1738} & 0.0607 \\
Mid (\( \theta \in [-1, 1] \)) & 251 & 0.1139 & 0.0080 \\
High (\( \theta > 1 \)) & 82 & 0.0111 & 0.0009 \\
\hline
\end{tabular}
\end{table}

\subsection{Main Results: Feedback Quality}
\subsubsection{Pairwise Comparison Result}
Figure~\ref{fig:ablation_feedback} presents the pairwise comparison results of PsyScore-AEF against strong baselines. \textbf{Actionability vs. General LLMs:} When compared to general-purpose models such as GPT-4o and Llama-3, PsyScore-AEF achieves win rates exceeding 90\% in the \textit{Actionability} dimension. This indicates that the Multi-Agent Fusion mechanism produces more concrete and executable feedback compared to standard generative models.

\textbf{Adaptivity via Psychometric Diagnosis:} Ablation analysis (Figure~\ref{fig:ablation_feedback}) demonstrates that removing the psychometric module results in a substantial reduction in both \textit{Adaptivity} and \textit{Tone}, with PsyScore retaining a win rate above 75\%. This confirms that conditioning feedback generation on latent ability ($\theta$) and discrimination ($a$) parameters is essential for aligning instructional scaffolding with the learner's ZPD.

\subsubsection{Simulation-based Revision Result} 

\begin{table*}[!h]
\centering
\caption{Expert Evaluation Results (Mean $\pm$ SD on a 1-5 Likert Scale).}
\label{tab:expert-eval}
\begin{tabular}{lccccc}
\hline
\textbf{Model} & \textbf{Accuracy} & \textbf{Actionability} & \textbf{Adaptivity} & \textbf{Specificity} & \textbf{Tone} \\
\hline
GPT-4o & 3.10 $\pm$ .45 & 2.88 $\pm$ .77 & 2.79 $\pm$ .53 & 2.55 $\pm$ .61 & 3.15 $\pm$ .63 \\
Llama-4 & 2.88 $\pm$ .52 & 2.42 $\pm$ .66 & 2.66 $\pm$ .68 & 2.34 $\pm$ .50 & 2.62 $\pm$ .69 \\
Qwen-3 & 3.13 $\pm$ .49 & 2.91 $\pm$ .77 & 2.91 $\pm$ .83 & 2.70 $\pm$ .75 & 3.11 $\pm$ .74 \\
PsyScore-No-IRT & 3.65 $\pm$ .60 & 4.03 $\pm$ .63 & 3.48 $\pm$ .61 & 3.84 $\pm$ .84 & 3.77 $\pm$ .67 \\
\textbf{PsyScore} & \textbf{4.01 $\pm$ .48} & \textbf{4.13 $\pm$ .68} & \textbf{4.21 $\pm$ .54} & \textbf{4.17 $\pm$ .70} & \textbf{4.25 $\pm$ .63} \\
\hline
\end{tabular}
\end{table*}

Table~\ref{tab:revision-gain} reports normalized revision gains by student proficiency. PsyScore achieves \textbf{17.38\%} gain for low-proficiency ($\theta < -1$) vs.\ $6.07\%$ for No-IRT, confirming targeted scaffolding for foundational deficits. Mid-proficiency gains are $11.39\%$ (vs.\ $0.80\%$), while high-proficiency gains are modest ($1.11\%$ vs.\ $0.09\%$) due to a ceiling effect. Ablation confirms that removing the psychometric layer ($\theta$) severely degrades gains, particularly for low-ability learners, validating $\theta$'s role in calibrating feedback to cognitive readiness and avoiding over-assistance.

The resulting expert ratings, together with comprehensive inter-rater reliability statistics, are reported in Table\ref{tab:expert-eval}. PsyScore significantly outperforms all LLM baselines (GPT-4o, Llama-4, Qwen-3) on every dimension ($p < .001$, paired $t$-test with Bonferroni correction). The ablation variant (\texttt{PsyScore-No-IRT}) exhibits marked declines in \textit{Adaptivity} and \textit{Accuracy} ($p < .05$), confirming that IRT-based student modeling is indispensable for generating feedback that is both contextually appropriate and diagnostically precise. Inter-rater reliability analysis indicates substantial agreement among the three experts: Fleiss' $\kappa$ reaches $0.733$ for Specificity and $0.690$ for Accuracy, while all pairwise weighted Cohen's $\kappa$ values exceed $0.60$. These metrics collectively validate the robustness and pedagogical validity of the expert assessments.

\section{Discussion and Conclusion}
\subsection{Discussion}
The IRT layer acts as a psychometric regularizer, constraining representations to improve robustness on high-variance traits like \textit{Voice} (Prompt~8). This transforms AES from summative scoring into formative diagnosis. Simulated revision confirms ZPD-aligned scaffolding: PsyScore achieves a \textbf{17.38\%} normalized gain for low-proficiency students ($\theta < -1$), far exceeding the baseline. Gains diminish for advanced learners due to a ceiling effect, underscoring that evaluation should prioritize long-term competency development over immediate text improvement.

\subsection{Conclusion}
PsyScore integrates psychometric calibration with ZPD-conditioned multi-agent feedback, advancing AES from static scoring to adaptive instructional scaffolding. Trait-adaptive IRT priors yield state-of-the-art scoring accuracy, while ability-aware feedback generation dynamically aligns support with learner proficiency, advocating a formative turn in educational AI.

\section*{Limitations}
\textbf{Computational Overhead.} The multi-agent consensus mechanism trades off inference latency for pedagogical reliability. While acceptable for asynchronous feedback, this limits real-time classroom deployment. Future compression via knowledge distillation could preserve diagnostic fidelity while enabling interactive responsiveness.

\textbf{Annotation Dependency.} The trait-adaptive scorer requires fine-grained analytic labels (e.g., \textit{Voice}, \textit{Organization}), which are scarce in operational assessment settings. This restricts immediate generalization to holistic-only corpora. Semi-supervised trait disentanglement offers a promising path toward label-efficient psychometric modeling.

\textbf{Ecological Validity of Simulation.} The simulated revision protocol captures \textit{competency gains} under idealized adherence, but cannot model motivational dynamics, epistemic trust, or cognitive fatigue that mediate real-world feedback uptake. Controlled classroom trials are necessary to validate whether scaffolding benefits persist under authentic instructional conditions.
\bibliography{custom}
\newpage
\appendix

\section{ASAP++ Dataset Details}
\label{sec:appendix-dataset}

The \textbf{ASAP++} dataset~\cite{mathias2018asap++} extends the original Automated Student Assessment Prize (ASAP) corpus by providing fine-grained, trait-specific scores in addition to holistic essay ratings. It comprises eight distinct writing prompts spanning three genres: argumentative (persuasive essays), source-based responses, and narrative storytelling. Table~\ref{tab:asap_dataset} summarizes the prompt-level statistics, including the number of essays, average length, and overall score range for each task.

Each essay is evaluated along multiple analytic dimensions, the set of which varies by prompt. Commonly assessed traits include \textit{Content}, \textit{Organization}, \textit{Word Choice}, \textit{Sentence Fluency}, \textit{Conventions}, \textit{Prompt Adherence}, \textit{Narrativity}, \textit{Language}, \textit{Style}, and \textit{Voice}. Score ranges for individual traits differ from holistic scores; for instance, Prompt~8 assigns holistic scores on a 0--60 scale while trait scores follow a 2--12 range. A detailed mapping of trait dimensions and their corresponding score intervals is provided in Table~\ref{tab:asap_dataset}.

\begin{table}[!htbp]
\centering
\caption{Basic statistics of the ASAP-AES dataset by prompt.}
\label{tab:asap_dataset}
\setlength{\tabcolsep}{2pt}

\begin{tabular}{ccccc}
\toprule
\textbf{Prompt} & \textbf{Type} & \textbf{Count} & \textbf{Length} & \textbf{Range} \\
\midrule
Prompt 1 & Argumentative & 1,785 & 350 & 2--12 \\
Prompt 2 & Argumentative & 1,800 & 350 & 1--6 \\
Prompt 3 & Response & 1,726 & 150 & 0--3 \\
Prompt 4 & Response & 1,772 & 150 & 0--3 \\
Prompt 5 & Response & 1,805 & 150 & 0--4 \\
Prompt 6 & Response & 1,800 & 150 & 0--4 \\
Prompt 7 & Narrative & 1,569 & 300 & 0--30 \\
Prompt 8 & Narrative & 723 & 650 & 0--60 \\
\bottomrule
\end{tabular}
\end{table}

\section{Initialization and Ensemble Strategy}
\label{sec:appendix-a}
Table \ref{tab:appendix_full_grid_search} presents a comprehensive grid search over IRT parameter initialization strategies. We evaluate nine combinations of discrimination ($a \in \{0.5, 1.0, 1.5\}$) and difficulty ($b \in \{1.0, 2.0, 3.0\}$) parameters across all writing traits and prompts using two metrics: Quadratic Weighted Kappa (QWK) and Pearson Correlation Coefficient (PCC). The results demonstrate that moderate initialization values ($a=1.0$, $b=2.0$) generally yield optimal performance, with consistent stability across different writing dimensions. The table provides empirical evidence supporting our parameter selection strategy for the diagnostic feedback system.

\begin{table*}[!htbp]
\vspace*{\fill}  
\centering
\fontfamily{ptm}\selectfont 
\setlength{\tabcolsep}{4.5pt}
\renewcommand{\arraystretch}{1.2}
\caption{The table reports metrics across all traits and prompts for nine different initialization combinations of discrimination ($a$) and difficulty ($b$).}
\label{tab:appendix_full_grid_search}
\begin{sideways} 
\centering
\begin{minipage}{\textheight} 
\begin{tabular}{@{}l@{}*{18}{c}@{}}
\toprule
\multirow{3}{*}{\textbf{Dimensions}} & 
\multicolumn{2}{c}{\textbf{A=0.5}} & 
\multicolumn{2}{c}{\textbf{A=0.5}} & 
\multicolumn{2}{c}{\textbf{A=0.5}} & 
\multicolumn{2}{c}{\textbf{A=1.0}} & 
\multicolumn{2}{c}{\textbf{A=1.0}} & 
\multicolumn{2}{c}{\textbf{A=1.0}} & 
\multicolumn{2}{c}{\textbf{A=1.5}} & 
\multicolumn{2}{c}{\textbf{A=1.5}} & 
\multicolumn{2}{c}{\textbf{A=1.5}} \\
& 
\multicolumn{2}{c}{\textbf{B=[-1, 1]}} & 
\multicolumn{2}{c}{\textbf{B=[-2, 2]}} & 
\multicolumn{2}{c}{\textbf{B=[-3, 3]}} & 
\multicolumn{2}{c}{\textbf{B=[-1, 1]}} & 
\multicolumn{2}{c}{\textbf{B=[-2, 2]}} & 
\multicolumn{2}{c}{\textbf{B=[-3, 3]}} & 
\multicolumn{2}{c}{\textbf{B=[-1, 1]}} & 
\multicolumn{2}{c}{\textbf{B=[-2, 2]}} & 
\multicolumn{2}{c}{\textbf{B=[-3, 3]}} \\
\cmidrule(lr){2-3}\cmidrule(lr){4-5}\cmidrule(lr){6-7}\cmidrule(lr){8-9}\cmidrule(lr){10-11}\cmidrule(lr){12-13}\cmidrule(lr){14-15}\cmidrule(lr){16-17}\cmidrule(lr){18-19}
& QWK & PCC & QWK & PCC & QWK & PCC & QWK & PCC & QWK & PCC & QWK & PCC & QWK & PCC & QWK & PCC & QWK & PCC \\
\midrule
\multicolumn{19}{c}{\textit{Trait-wise Performance}} \\
\midrule
Overall          & 0.775 & 0.782 & 0.774 & 0.784 & 0.783 & 0.785 & 0.775 & 0.782 & 0.785 & 0.786 & 0.781 & 0.784 & 0.779 & 0.782 & 0.778 & 0.781 & 0.780 & 0.782 \\
Content          & 0.738 & 0.755 & 0.743 & 0.753 & 0.738 & 0.751 & 0.738 & 0.755 & 0.752 & 0.759 & 0.750 & 0.755 & 0.754 & 0.759 & 0.754 & 0.758 & 0.747 & 0.755 \\
Organization     & 0.709 & 0.721 & 0.710 & 0.714 & 0.702 & 0.711 & 0.709 & 0.721 & 0.696 & 0.710 & 0.696 & 0.700 & 0.692 & 0.700 & 0.705 & 0.710 & 0.691 & 0.696 \\
Word Choice      & 0.693 & 0.697 & 0.694 & 0.696 & 0.704 & 0.715 & 0.693 & 0.697 & 0.715 & 0.713 & 0.716 & 0.717 & 0.710 & 0.717 & 0.708 & 0.717 & 0.701 & 0.705 \\
Sentence Fluency & 0.717 & 0.720 & 0.703 & 0.721 & 0.691 & 0.700 & 0.717 & 0.720 & 0.700 & 0.704 & 0.703 & 0.708 & 0.708 & 0.711 & 0.697 & 0.713 & 0.705 & 0.714 \\
Conventions      & 0.714 & 0.731 & 0.716 & 0.734 & 0.717 & 0.729 & 0.714 & 0.731 & 0.709 & 0.719 & 0.713 & 0.723 & 0.724 & 0.730 & 0.725 & 0.738 & 0.702 & 0.708 \\
Prompt Adherence & 0.740 & 0.760 & 0.739 & 0.751 & 0.745 & 0.754 & 0.740 & 0.760 & 0.754 & 0.758 & 0.746 & 0.751 & 0.743 & 0.765 & 0.751 & 0.761 & 0.743 & 0.752 \\
Narrativity      & 0.720 & 0.732 & 0.722 & 0.740 & 0.720 & 0.734 & 0.720 & 0.732 & 0.706 & 0.722 & 0.726 & 0.737 & 0.731 & 0.750 & 0.727 & 0.734 & 0.722 & 0.740 \\
Language         & 0.689 & 0.714 & 0.690 & 0.709 & 0.704 & 0.718 & 0.689 & 0.714 & 0.702 & 0.710 & 0.697 & 0.708 & 0.668 & 0.687 & 0.700 & 0.715 & 0.696 & 0.718 \\
Style            & 0.641 & 0.686 & 0.693 & 0.726 & 0.668 & 0.682 & 0.641 & 0.686 & 0.684 & 0.703 & 0.679 & 0.702 & 0.685 & 0.712 & 0.681 & 0.686 & 0.680 & 0.705 \\
Voice            & 0.701 & 0.692 & 0.738 & 0.733 & 0.742 & 0.756 & 0.701 & 0.692 & 0.711 & 0.710 & 0.723 & 0.738 & 0.705 & 0.732 & 0.641 & 0.697 & 0.727 & 0.737 \\
Trait Avg        & 0.712 & 0.727 & 0.720 & 0.733 & 0.720 & 0.730 & 0.712 & 0.727 & 0.719 & 0.727 & 0.721 & 0.729 & 0.718 & 0.731 & 0.715 & 0.728 & 0.718 & 0.728 \\
\midrule
\multicolumn{19}{c}{\textit{Prompt-wise Performance}} \\
\midrule
Prompt 1         & 0.733 & 0.748 & 0.732 & 0.749 & 0.743 & 0.754 & 0.733 & 0.748 & 0.733 & 0.742 & 0.731 & 0.739 & 0.743 & 0.749 & 0.741 & 0.747 & 0.724 & 0.732 \\
Prompt 2         & 0.712 & 0.720 & 0.707 & 0.713 & 0.722 & 0.727 & 0.712 & 0.720 & 0.703 & 0.714 & 0.717 & 0.721 & 0.711 & 0.718 & 0.704 & 0.717 & 0.697 & 0.706 \\
Prompt 3         & 0.707 & 0.731 & 0.724 & 0.740 & 0.705 & 0.720 & 0.707 & 0.731 & 0.720 & 0.727 & 0.730 & 0.739 & 0.724 & 0.732 & 0.712 & 0.724 & 0.722 & 0.736 \\
Prompt 4         & 0.759 & 0.775 & 0.763 & 0.777 & 0.764 & 0.770 & 0.759 & 0.775 & 0.764 & 0.775 & 0.762 & 0.769 & 0.762 & 0.773 & 0.772 & 0.784 & 0.768 & 0.771 \\
Prompt 5         & 0.721 & 0.739 & 0.705 & 0.725 & 0.712 & 0.725 & 0.721 & 0.739 & 0.727 & 0.733 & 0.711 & 0.718 & 0.712 & 0.731 & 0.733 & 0.735 & 0.712 & 0.730 \\
Prompt 6         & 0.741 & 0.764 & 0.752 & 0.766 & 0.765 & 0.772 & 0.741 & 0.764 & 0.760 & 0.767 & 0.757 & 0.767 & 0.736 & 0.761 & 0.761 & 0.771 & 0.758 & 0.771 \\
Prompt 7         & 0.743 & 0.761 & 0.753 & 0.766 & 0.744 & 0.749 & 0.743 & 0.761 & 0.753 & 0.758 & 0.752 & 0.762 & 0.752 & 0.761 & 0.758 & 0.762 & 0.749 & 0.757 \\
Prompt 8         & 0.706 & 0.700 & 0.707 & 0.707 & 0.697 & 0.712 & 0.706 & 0.700 & 0.709 & 0.709 & 0.710 & 0.711 & 0.712 & 0.713 & 0.697 & 0.709 & 0.713 & 0.715 \\
Prompt Avg       & 0.728 & 0.742 & 0.730 & 0.743 & 0.731 & 0.741 & 0.728 & 0.742 & 0.734 & 0.741 & 0.734 & 0.741 & 0.732 & 0.742 & 0.735 & 0.744 & 0.730 & 0.740 \\
\bottomrule
\end{tabular}
\end{minipage}
\end{sideways}
\vspace*{\fill}
\end{table*}


\section{ZPD-Aligned Strategy Mapping}
Table~\ref{tab:zpd_strategy} presents our Zone of Proximal Development (ZPD) aligned pedagogical strategy mapping that dynamically selects teaching approaches based on the student's latent ability parameter ($\theta$) estimated through our IRT model. The system implements three distinct instructional strategies: explicit directive instruction for struggling learners ($\theta < -1.0$), targeted scaffolding for developing learners ($-1.0 \leq \theta \leq 1.0$), and intellectual challenge for advanced learners ($\theta > 1.0$). This adaptive strategy selection forms the foundation of our diagnostic feedback generation pipeline, ensuring pedagogical appropriateness for each student's proficiency level.
\\
\label{sec:appendix-b}
\begin{table}[!htbp]
\centering
\caption{Zone of Proximal Development (ZPD) aligned pedagogical strategy mapping based on student latent ability ($\theta$)}
\label{tab:zpd_strategy}
\begin{tabular}{p{0.18\columnwidth}p{0.23\columnwidth}p{0.5\columnwidth}}
\toprule
\textbf{$\theta$ Range} & \textbf{Student Proficiency} & \textbf{Pedagogical Strategy} \\
\midrule
$\theta < -1.0$ & Struggling Learner & 
\begin{minipage}[t]{0.5\columnwidth}
\texttt{Strategy: EXPLICIT INSTRUCTION (Directive)} \\
\texttt{The student struggles with fundamentals.} \\
\texttt{Action: Provide direct corrections for grammar and mechanics.} \\
\texttt{Explanation: Keep explanations concise and rule-based.} \\
\texttt{Goal: Fix errors that block communication.}
\end{minipage} \\
\midrule
$-1.0 \leq \theta \leq 1.0$ & Developing Learner & 
\begin{minipage}[t]{0.5\columnwidth}
\texttt{Strategy: TARGETED SCAFFOLDING} \\
\texttt{The student has emerging competence.} \\
\texttt{Action: Balance praise with 1-2 key areas for improvement.} \\
\texttt{Goal: Guide the student to the next proficiency level (ZPD).}
\end{minipage} \\
\midrule
$\theta > 1.0$ & Advanced Learner & 
\begin{minipage}[t]{0.5\columnwidth}
\texttt{Strategy: INTELLECTUAL CHALLENGE (Facilitative)} \\
\texttt{The student shows strong command.} \\
\texttt{Action: Focus on rhetorical analysis, logic flow, and voice.} \\
\texttt{Explanation: Use open-ended questions to trigger self-reflection.} \\
\texttt{Goal: Refine style and depth of argument.}
\end{minipage} \\
\bottomrule
\end{tabular}
\end{table}

\section{Multi-Agent Consensus and Debiasing}
\label{sec:appendix-c}
Table~\ref{tab:prompts} details the three-stage prompting architecture used in our diagnostic feedback generation pipeline. The first stage generates personalized pedagogical instructions based on the student's latent ability parameter (theta) and item response theory metrics. The second stage employs multiple expert writing tutors to generate initial feedback drafts following these instructions. The final stage uses an expert editor model to synthesize the best elements from all drafts into a cohesive, pedagogically appropriate final feedback report. This multi-stage approach ensures both diagnostic precision and pedagogical effectiveness in the generated feedback.
\begin{table}[!htbp]
\centering
\caption{Prompt templates used at different stages of the diagnostic feedback generation pipeline}
\label{tab:prompts}
\begin{tabular}{p{0.3\columnwidth}p{0.65\columnwidth}}
\toprule
\textbf{Stage} & \textbf{Prompt Template} \\
\midrule
Diagnostic Instruction Generation & 
\begin{minipage}[t]{0.65\columnwidth}
\texttt{1. PEDAGOGICAL STRATEGY PROTOCOL} \\
\texttt{> Student Latent Ability (Theta): <theta\_value>} \\
\texttt{> Required Approach: <strategy\_description>} \\
\texttt{CRITICAL INSTRUCTION: Must adopt defined 'Tone' and execute 'Action'.} \\
\texttt{2. FEEDBACK PRIORITY MAP} \\
\texttt{- <trait> (a=<value>): [CRITICAL FOCUS/Standard/Secondary]} \\
\texttt{3. DIAGNOSTIC INTERVENTION POINTS} \\
\texttt{- <trait>: <status> (Score: <score>)}
\end{minipage} \\
\midrule
Initial Feedback Generation & 
\begin{minipage}[t]{0.65\columnwidth}
\texttt{Role: Expert Writing Tutor implementing ``Differentiated Instruction''} \\
\texttt{[STUDENT ESSAY] <full essay text>} \\
\texttt{[PEDAGOGICAL DIAGNOSIS \& INSTRUCTIONS] <diagnostic instructions>} \\
\texttt{Task: Write personalized feedback adhering to the defined Strategy.} \\
\texttt{Output: Return feedback content directly.}
\end{minipage} \\
\midrule
Expert Synthesis & 
\begin{minipage}[t]{0.65\columnwidth}
\texttt{Role: Chief Editor of Educational Feedback} \\
\texttt{[TRUTH - STUDENT PROFILE]} \\
\texttt{- Latent Ability: <value>} \\
\texttt{- Required Strategy: <strategy\_desc>} \\
\texttt{[DRAFT FEEDBACK FROM TUTORS] <multiple candidate drafts>} \\
\texttt{Task: Combine best insights into ONE FINAL PERFECT FEEDBACK REPORT} \\
\texttt{Fusion Rules:} \\
\texttt{1. Filter by Strategy: Ignore parts violating required strategy} \\
\texttt{2. ......} \\
\texttt{Output: Return ONLY the final synthesized feedback text.}
\end{minipage} \\
\bottomrule
\end{tabular}
\end{table}

\section{Full Statistical Significance Results}
\label{app:full_statistical_table}
Table~\ref{tab:performance_comparison} presents the complete per-trait scoring performance of PsyScore-MAES compared against the BERT baseline across all eight prompts in the ASAP++ dataset. For each prompt and each evaluated writing trait, we report the mean Quadratic Weighted Kappa (QWK) and its standard deviation over five cross-validation folds, the absolute difference in mean QWK between the two models (Diff.), the exact one-tailed \(p\)-value derived from the Wilcoxon Signed-Rank Test, and an asterisk denoting statistical significance at the \(p < 0.05\) level. Traits without a defined score range in a given prompt are omitted from the corresponding row group.

\begin{table*}[!htbp]
    \centering
    \caption{Performance comparison between PsyScore-MAES and the BERT baseline across prompts and traits.}
    \label{tab:performance_comparison}
    \begin{tabularx}{\textwidth}{c c c c c c c}
        \toprule
        \textbf{Prompt} & \textbf{Trait} & \textbf{PsyScore-MAES} & \textbf{BERT} & \textbf{Diff.} & \textbf{$p$-value} & \textbf{Sig.} \\
        \midrule
        1 & Overall & 0.831 $\pm$ 0.007 & 0.794 $\pm$ 0.013 & 0.037 & 0.031 & * \\
          & Content & 0.730 $\pm$ 0.012 & 0.716 $\pm$ 0.008 & 0.014 & 0.031 & * \\
          & Organization & 0.696 $\pm$ 0.013 & 0.669 $\pm$ 0.016 & 0.028 & 0.031 & * \\   
          & Conventions & 0.706 $\pm$ 0.018 & 0.704 $\pm$ 0.010 & -0.001 & 0.500 &   \\
          & Word Choice & 0.716 $\pm$ 0.014 & 0.705 $\pm$ 0.018 & 0.011 & 0.094 &   \\
          & Sentence Fluency & 0.689 $\pm$ 0.018 & 0.702 $\pm$ 0.008 & -0.013 & 0.844 &   \\
        \midrule
        2 & Overall & 0.680 $\pm$ 0.020 & 0.644 $\pm$ 0.021 & 0.036 & 0.063 &   \\
          & Content & 0.688 $\pm$ 0.019 & 0.659 $\pm$ 0.010 & 0.030 & 0.031 & * \\
          & Organization & 0.689 $\pm$ 0.018 & 0.669 $\pm$ 0.014 & 0.020 & 0.031 & * \\
          & Conventions & 0.721 $\pm$ 0.017 & 0.688 $\pm$ 0.020 & 0.033 & 0.031 & * \\
          & Word Choice & 0.743 $\pm$ 0.011 & 0.708 $\pm$ 0.010 & 0.035 & 0.031 & * \\
          & Sentence Fluency & 0.718 $\pm$ 0.008 & 0.691 $\pm$ 0.018 & 0.027 & 0.063 &   \\
        \midrule
        3 & Overall & 0.707 $\pm$ 0.011 & 0.713 $\pm$ 0.014 & -0.006 & 0.844 &   \\
          & Content & 0.715 $\pm$ 0.012 & 0.695 $\pm$ 0.013 & 0.020 & 0.031 & * \\
          & Prompt Adherence & 0.740 $\pm$ 0.007 & 0.727 $\pm$ 0.008 & 0.013 & 0.031 & * \\
          & Language & 0.689 $\pm$ 0.020 & 0.684 $\pm$ 0.018 & 0.005 & 0.406 &   \\
          & Narrativity & 0.743 $\pm$ 0.012 & 0.741 $\pm$ 0.018 & 0.002 & 0.313 &   \\
        \midrule
        4 & Overall & 0.779 $\pm$ 0.003 & 0.774 $\pm$ 0.012 & 0.005 & 0.219 &   \\
          & Content & 0.772 $\pm$ 0.007 & 0.758 $\pm$ 0.023 & 0.015 & 0.406 &  \\
          & Prompt Adherence & 0.766 $\pm$ 0.012 & 0.755 $\pm$ 0.012 & 0.011 & 0.156 & \\
          & Language & 0.708 $\pm$ 0.012 & 0.677 $\pm$ 0.017 & 0.031 & 0.031 &  * \\
          & Narrativity & 0.782 $\pm$ 0.008 & 0.753 $\pm$ 0.014 & 0.029 & 0.031 &  * \\
        \midrule
        5 & Overall & 0.815 $\pm$ 0.006 & 0.797 $\pm$ 0.007 & 0.018 & 0.031 & * \\
          & Content & 0.738 $\pm$ 0.014 & 0.720 $\pm$ 0.015 & 0.018 & 0.063 & \\
          & Prompt Adherence & 0.700 $\pm$ 0.032 & 0.692 $\pm$ 0.008 & 0.004 & 0.313 & \\
          & Language & 0.706 $\pm$ 0.004 & 0.684 $\pm$ 0.015 & 0.021 & 0.063 &   \\
          & Narrativity & 0.684 $\pm$ 0.001 & 0.654 $\pm$ 0.004 & 0.030 & 0.031 &  * \\
        \midrule
        6 & Overall & 0.808 $\pm$ 0.010 & 0.801 $\pm$ 0.014 & 0.008 & 0.156 &   \\
          & Content & 0.824 $\pm$ 0.011 & 0.800 $\pm$ 0.015 & 0.024 & 0.031 & * \\
          & Prompt Adherence & 0.783 $\pm$ 0.015 & 0.766 $\pm$ 0.016 & 0.016 & 0.063 & \\
          & Language & 0.673 $\pm$ 0.018 & 0.682 $\pm$ 0.031 & -0.009 & 0.781 &   \\
          & Narrativity & 0.708 $\pm$ 0.022 & 0.674 $\pm$ 0.012 & 0.034 & 0.063 &   \\
        \midrule
        7 & Overall & 0.829 $\pm$ 0.006 & 0.779 $\pm$ 0.011 & 0.050 & 0.031 & * \\
          & Content & 0.858 $\pm$ 0.007 & 0.708 $\pm$ 0.017 & 0.149 & 0.031 & * \\
          & Organization & 0.679 $\pm$ 0.019 & 0.539 $\pm$ 0.031 & 0.140 & 0.031 & * \\
          & Conventions & 0.710 $\pm$ 0.012 & 0.456 $\pm$ 0.029 & 0.253 & 0.031 & * \\
          & Style & 0.664 $\pm$ 0.021 & 0.419 $\pm$ 0.022 & 0.245 & 0.031 & * \\
        \midrule
        8 & Overall & 0.791 $\pm$ 0.015 & 0.667 $\pm$ 0.031 & 0.125 & 0.031 & * \\
          & Content & 0.666 $\pm$ 0.019 & 0.505 $\pm$ 0.052 & 0.161 & 0.031 & * \\
          & Organization & 0.704 $\pm$ 0.017 & 0.532 $\pm$ 0.029 & 0.172 & 0.031 & * \\
          & Conventions & 0.697 $\pm$ 0.043 & 0.535 $\pm$ 0.033 & 0.162 & 0.031 & * \\
          & Word Choice & 0.674 $\pm$ 0.022 & 0.409 $\pm$ 0.058 & 0.265 & 0.031 & * \\
          & Sentence Fluency & 0.678 $\pm$ 0.038 & 0.511 $\pm$ 0.029 & 0.167 & 0.031 & * \\
          & Voice & 0.721 $\pm$ 0.017 & 0.484 $\pm$ 0.031 & 0.237 & 0.031 & * \\
        \bottomrule
    \end{tabularx}
\end{table*}

\section{Preference Compare Evaluation}
\label{sec:appendix-d}
Table~\ref{tab:preference_comparison_prompt_v2} presents the prompt template used for human-aligned preference evaluation between feedback variants. The prompt instructs an LLM to act as a pedagogical expert evaluating two feedback versions across three critical educational dimensions: specificity (citing concrete examples from the student essay), actionability (providing clear revision guidance), and pedagogical value (supporting student learning). The structured JSON output format ensures consistent, machine-readable evaluation results. This multi-dimensional assessment framework allows us to quantitatively compare the pedagogical quality of different feedback generation approaches while maintaining alignment with educational best practices.
\begin{table}[!htbp]
\centering
\caption{Prompt template used for multi-dimensional preference comparison between feedback variants}
\label{tab:preference_comparison_prompt_v2}
\begin{tabular}{p{0.95\columnwidth}}
\toprule
\textbf{Evaluation Prompt Template} \\
\midrule
\begin{minipage}[t]{0.95\columnwidth}
\texttt{You are an expert Pedagogical Evaluator.} \\
\\
\texttt{[STUDENT ESSAY]:} \\
\texttt{<essay content (smart truncated)>} \\
\\
\texttt{[FEEDBACK A]:} \\
\texttt{<feedback variant A>} \\
\\
\texttt{[FEEDBACK B]:} \\
\texttt{<feedback variant B>} \\
\\
\texttt{[TASK]:} \\
\texttt{Compare Feedback A and Feedback B along THREE distinct dimensions.} \\
\texttt{For each dimension, choose a winner (``A'', ``B'', or ``Tie'') and provide a brief reason.} \\
\\
\texttt{1. \textbf{Specificity}: Does the feedback cite specific text from the essay?} \\
\texttt{2. \textbf{Actionability}: Does it give clear instructions on HOW to revise?} \\
\texttt{3. \textbf{Pedagogical\_Value}: Is the feedback supportive and conducive to learning?} \\
\\
\texttt{[OUTPUT FORMAT]:} \\
\texttt{Return ONLY a valid JSON object with these exact keys.} \\
\texttt{\{} \\
\texttt{    ``specificity'': \{} \\
\texttt{        ``winner'': ``A'' or ``B'' or ``Tie'', } \\
\texttt{        ``reason'': ``explanation''} \\
\texttt{    \},} \\
\texttt{    ``actionability'': \{} \\
\texttt{        ``winner'': ``A'' or ``B'' or ``Tie``, } \\
\texttt{        ``reason'': ``explanation''} \\
\texttt{    \},} \\
\texttt{    ``pedagogical'': \{} \\
\texttt{        ``winner'': ``A'' or ``B'' or ``Tie'', } \\
\texttt{        ``reason'': ``explanation''} \\
\texttt{    \},} \\
\texttt{    ``overall\_preference'': ``A'' or ``B'' or ``Tie''} \\
\texttt{\}}
\end{minipage} \\
\bottomrule
\end{tabular}
\end{table}

\section{Simulated Scaffolding Assessment}
\label{sec:appendix-e}
Table~\ref{tab:simulation_prompts} outlines the dual-role prompt templates used in our simulated pedagogical evaluation framework. The first template simulates a student who revises their essay strictly according to provided feedback, with explicit constraints ensuring revisions remain faithful to the feedback content. The second template simulates a professional essay grader who evaluates essay quality on a standardized 1.0-6.0 scale across multiple dimensions (Organization, Content, Grammar, and Style) with structured JSON output. This simulation framework allows us to quantitatively measure feedback effectiveness by tracking score improvements between original and revised essays, providing an automated yet pedagogically grounded evaluation of feedback quality.
\begin{table}[!htbp]
\centering
\caption{Prompt templates used in the simulated pedagogical evaluation framework}
\label{tab:simulation_prompts}
\begin{tabular}{p{0.2\columnwidth}p{0.75\columnwidth}}
\toprule
\textbf{Role} & \textbf{Prompt Template} \\
\midrule
Student Simulator & 
\begin{minipage}[t]{0.75\columnwidth}
\texttt{You are a student revising your essay based on a teacher's feedback.} \\
\texttt{[YOUR ORIGINAL ESSAY]:} \\
\texttt{<original essay text>} \\
\texttt{[TEACHER'S FEEDBACK]:} \\
\texttt{``<feedback content>''} \\
\texttt{[TASK]:} \\
\texttt{Revise your essay to improve its quality.} \\
\texttt{**CRITICAL RULES**:} \\
\texttt{1. You must **ONLY** make changes that are suggested or implied by the feedback.} \\
\texttt{2. If the feedback is specific (e.g., ``fix the intro''), focus on that.} \\
\texttt{3. If the feedback is generic (e.g., ``write better''), try your best but do not rewrite the whole essay from scratch.} \\
\texttt{4. Do not add conversational text. Output ONLY the revised essay.}
\end{minipage} \\
\midrule
Essay Grader & 
\begin{minipage}[t]{0.75\columnwidth}
\texttt{You are a professional Essay Grader.} \\
\texttt{[ESSAY]:} \\
\texttt{<essay text>} \\
\texttt{[TASK]:} \\
\texttt{Rate this essay on a scale of 1.0 to 6.0 (increments of 0.5 allowed).} \\
\texttt{Assess: Organization, Content, Grammar, and Style.} \\
\texttt{- 1.0 = Very Poor} \\
\texttt{- 6.0 = Excellent} \\
\texttt{[OUTPUT FORMAT]:} \\
\texttt{Return ONLY a valid JSON object.} \\
\texttt{Do not simply copy the example values; determine the score based strictly on the essay quality.} \\
\texttt{Example Structure:} \\
\texttt{\{} \\
\texttt{    ``score'': <FLOAT\_BETWEEN\_1\_AND\_6>,} \\
\texttt{    ``reason'': ``<YOUR\_SHORT\_JUSTIFICATION>''} \\
\texttt{\}}
\end{minipage} \\
\bottomrule
\end{tabular}
\end{table}

\section{Human Expert Rubric}
\label{sec:appendix-f}
To address the limitations of automated evaluation in capturing pedagogical context and to validate the external validity of the experimental findings, we invited three experts with extensive experience in English language instruction to conduct a fine-grained human evaluation. The assessment employed a five-point Likert scale, with detailed criteria provided in Table~\ref{tab:human_expert_rubric}.
\clearpage
\begin{table*}[!htbp]
\centering
\caption{Human Expert Rubric for Writing Feedback Quality Assessment}
\label{tab:human_expert_rubric}
\footnotesize 
\renewcommand{\arraystretch}{1.0} 
\begin{tabularx}{\textwidth}{
    >{\raggedright\arraybackslash}p{1.0cm} 
    >{\raggedright\arraybackslash}p{1.2cm} 
    >{\raggedright\arraybackslash}p{1.5cm} 
    *{5}{X} 
}
\toprule
\textbf{Dim} & \textbf{Theoretical Grounding} & \textbf{Core Evaluation Focus} & \textbf{1 (Poor)} & \textbf{2 (Fair)} & \textbf{3 (Moderate)} & \textbf{4 (Good)} & \textbf{5 (Excellent)} \\ \midrule

1. Adaptivity 
& Zone of Proximal Development \cite{vygotsky1978mind}
& Does the depth and complexity of the feedback align with the learner's current cognitive level and ZPD? 
& Severe mismatch. Feedback is either far too difficult or too trivial , completely ignoring individual proficiency. 
& Low alignment. Feedback feels rigid and lacks meaningful personalization for the specific learner. 
& Moderate. No obvious logical errors, but exhibits a ``generic template'' style; lacks targeted adaptation. 
& Good alignment. Dynamically adjusts semantic depth or linguistic complexity based on inferred student ability. 
& Perfect alignment. Precisely targets the ZPD. Provides scaffolding for low-proficiency learners and facilitative challenges for advanced learners. \\ \midrule

2. Actionability 
& Feedforward Theory \cite{hattie2007power}
& Does the feedback provide clear revision paths or scaffolding that the learner can immediately act upon? 
& Not actionable. Offers only vague criticism with no direction for change; advice is confusing. 
& Vague instructions. Suggestions are overly abstract, leaving the student uncertain where to begin. 
& Acceptable direction. Indicates a general area for revision but lacks concrete steps or methodological support. 
& Clear instructions. Explicitly identifies a revision path that the student can follow independently, though specific examples may be absent. 
& Fully scaffolded. Provides actionable directives alongside revision strategies, worked examples, or stepwise reasoning guidance. \\ \midrule

3. Specificity 
& Cognitive Load Theory \cite{sweller1988cognitive} 
& Does the feedback cite specific textual evidence rather than offering vague evaluations that increase cognitive load? 
& Completely detached. Exhibits hallucination, referencing non-existent content, or is entirely off-topic. 
& Generic platitudes. Relies on boilerplate phrases with no connection to the actual text. 
& Slight relevance. Mentions the general topic or isolated keywords but fails to engage with textual details. 
& Concrete citation. Precisely references specific paragraphs, sentences, or phrases in the essay for targeted commentary. 
& Deep insight. Functions like a microscope, accurately citing original text and analyzing underlying logical or rhetorical issues with definitive evidence. \\ \midrule

4. Accuracy 
& Pedagogical Content Knowledge \cite{shulman1986those}
& Are diagnostic information, linguistic corrections, and factual statements objectively correct and free of misleading content? 
& Severely erroneous. Provides incorrect knowledge, misidentifies correct usage as error, or introduces factual mistakes. 
& Obvious mistakes. Contains clear grammatical misjudgments or factual inaccuracies in the diagnosis. 
& Generally accurate. No fundamental errors, but may overlook deeper logical issues, or suggestions are debatable. 
& Professionally accurate. Diagnostic conclusions are precise; instructional suggestions fully align with linguistic conventions and norms. 
& Exemplary precision. Beyond surface correction, accurately identifies subtle issues such as pragmatic missteps or logical inconsistencies. \\ \midrule

5. Supportive Tone 
& Affective Filter Hypothesis \cite{krashen1982principles}
& Is the wording empathetic, aiming to reduce learner anxiety while stimulating motivation to revise? 
& Negative and cold. Robotic or condescending tone; feedback consists solely of negative criticism. 
& Lacks warmth. Reads like a machine-generated error log; purely task-oriented with no humanistic encouragement. 
& Polite and objective. Neutral in tone; acceptable to the learner but uninspiring and unlikely to boost motivation. 
& Positively encouraging. Employs a constructive approach, fostering a positive feedback loop. 
& Highly inspiring. Demonstrates empathy and adopts a collaborative stance that cultivates a growth mindset and strong revision willingness. \\ 
\bottomrule
\end{tabularx}
\end{table*}

\clearpage
\end{document}